%% file: main.tex
\documentclass{article} %
\usepackage[T1]{fontenc}
\usepackage[utf8]{inputenc}
\usepackage{viola,times}
\input{math_commands.tex}

\usepackage{hyperref}
\hypersetup{pdftitle={VioLA: Learning Generalist Humanoid Control Policies from Human Data},
  pdfauthor={Mert Albaba, Jens Beißwenger, Anna Manasyan, Daniel Marta, Michael J. Black, Wieland Brendel, Andreas Krause, Georg Martius, Martin Riedmiller}}
\usepackage{url}
\usepackage{graphicx}
\usepackage{booktabs}
\newcommand{\methodname}{VioLA}

\usepackage{wrapfig}

\title{VioLA: Learning Generalist Humanoid \\ Control Policies from Human Data}

\author{%
  \parbox[t]{\dimexpr\linewidth-2\tabcolsep}{\centering\normalfont
    \textbf{Mert Albaba}\thanks{Equal contribution}$^{\hspace{0.15cm}1,2,3}$ \hspace{0.25cm}
    \textbf{Jens Bei\ss{}wenger}$^{*3,5}$ \hspace{0.25cm}
    \textbf{Anna Manasyan}$^{*3,4}$ \hspace{0.25cm}
    \textbf{Daniel Marta}$^{2}$ \\[0.8ex]
    \makebox[\linewidth][c]{%
      \textbf{Michael J. Black}$^{3}$
      \textbf{Wieland Brendel}$^{3,5}$
      \textbf{Andreas Krause}$^{2}$
      \textbf{Georg Martius}$^{3,4}$
      \textbf{Martin Riedmiller}$^{1}$%
    } \\[1.2ex]
    $^{1}$Vesoma \hspace{0.25cm} $^{2}$ETH Zürich \hspace{0.25cm} $^{3}$MPI-IS \hspace{0.25cm}
    $^{4}$University of Tuebingen \hspace{0.25cm} $^{5}$ELLIS Institute \\[0.8ex]
    \href{https://viola.is.tue.mpg.de}{viola.is.tue.mpg.de}
  }%
}

\begin{document}

\maketitle

\begin{abstract}
Teaching a humanoid to follow instructions with its whole body runs into two obstacles. Its action space is large and tightly coupled: legs, arms, and fingers must move together while the robot keeps its balance, which makes joint-level actions hard to learn. And humanoid demonstrations are scarce, so current humanoid generalist policies do not follow new instructions out of the box and are fine-tuned on teleoperated demonstrations of each task before deployment. Human demonstrations exist in far larger numbers, but a person's motion is not a robot command. We remove both obstacles by changing what the generalist policy predicts. We introduce VioLA, a generalist humanoid policy that predicts body and hand motion latents instead of joint commands. A pretrained body- and hand-controller execute these latents on the robot. Their corresponding motion encoders map human and robot motion into the same latent spaces. A human recording is therefore labeled in the policy's action space, and the training demonstration pool contains 140.6 million frames, 93.2\% of them human. As a result, VioLA follows locomotion instructions on the real robot zero-shot, without task-specific fine-tuning, reaching 100\% success where GR00T N1.7 and $\Psi_0$ reach 16.7\% and 0\%, respectively. It also reaches 88.6\% manipulation success without task-specific fine-tuning. The same approach works across two VLA and one world-action model backbones. A generalist policy trained on human demonstrations alone performs locomotion tasks on the real robot zero-shot. Code and checkpoints will be released.
\vspace{-7pt}
\end{abstract}

\begin{figure}[!htb]
\centering
\includegraphics[width=1\linewidth]{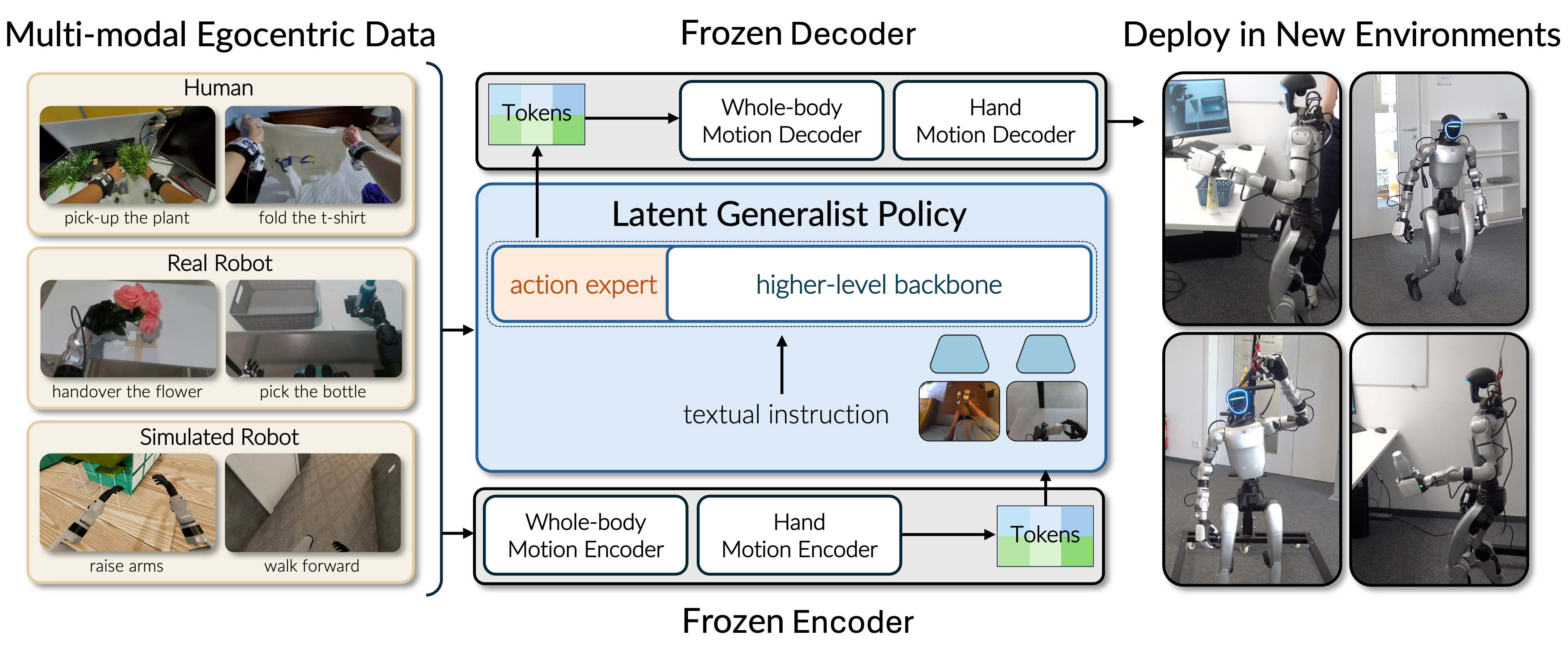}
\caption{\textbf{\methodname{}.} Human and robot motion encoders produce targets in shared body and hand latent spaces. The generalist policy predicts motion latents from vision, language, and proprioception; frozen robot controllers execute its predictions. Annotated human motion therefore supervises the policy in the same action representation as robot motion.}
\label{fig:teaser}
\end{figure}

\input{sections/intro}

\input{sections/related_works}

\input{sections/method}
\input{sections/experiments}
\input{sections/conclusion}

\section*{Acknowledgments}
Mert Albaba conducted this work during his internship at Vesoma. The authors thank our colleagues at Vesoma for their collaboration and for the many thoughtful discussions that shaped this work.\\
Daniel Marta was supported by the Knut and Alice Wallenberg Foundation.\\
Jens Bei\ss{}wenger was supported by the Federal Ministry of Research, Technology and Space of Germany (BMFTR, formerly BMBF) under grant no.~01IS24085C (OPENHAFM). \\
Georg Martius is a member of the Machine Learning Cluster of Excellence, EXC number 2064/1 -- Project number 390727645.
This work was supported by the German Federal Ministry of Education and Research (BMBF): Tübingen AI Center, FKZ: 01IS18039A. This work was supported by the ERC - 101045454 REAL-RL.

\input{main.bbl}
\clearpage
\appendix
{\Large\sc \methodname{}: Learning Generalist Humanoid\\ Control Policies from Human Data\par}
\vspace{0.5em}
{\large\sc Supplementary Material\par}

\input{sections/implementation}
\end{document}

%% file: math_commands.tex
\usepackage{amsmath,amsfonts,bm}

\def\eqref#1{equation~\ref{#1}}

\def\1{\bm{1}}

\DeclareMathAlphabet{\mathsfit}{\encodingdefault}{\sfdefault}{m}{sl}
\SetMathAlphabet{\mathsfit}{bold}{\encodingdefault}{\sfdefault}{bx}{n}

%% file: sections/intro.tex
\section{Introduction}
\label{sec:intro}
Consider telling a humanoid robot to walk to a table and close a laptop. The robot must find the table, walk to it, reach for the laptop, and close its screen without losing its balance. Vision-language-action models (VLAs) and world-action models (WAMs) learn to follow instructions from demonstrations, with substantial progress on robot arms ~\citep{rt2,openvla,pi05,dreamzero}. What keeps a similar humanoid generalist from following such instructions? Learning these behaviors on a humanoid presents two difficulties: coordinating the whole body and obtaining enough demonstrations to learn from.

The first obstacle is what the policy must predict. Whole-body control requires the legs, torso, arms, and fingers to move together. Reaching shifts the robot's weight, and stepping changes where its hands can reach. A generalist policy that predicts joint commands  must learn how to coordinate these movements while also learning which movement the instruction requires. Whole-body controllers already solve the coordination half. Given a desired motion, a pretrained whole-body controller computes joint commands to follow the motion while maintaining balance ~\citep{omnih2o,hover,sonic}. Recent latent motion controllers accept the desired motion as a compact learned vector, or motion latent, produced by an encoder ~\citep{ase,latent_space_priors,moconvq,sonic}. 

The second obstacle is data.  The robot demonstrations available for learning diverse tasks are limited. Humanoid teleoperation requires a robot and an operator, making it expensive to record many tasks in many environments. Existing humanoid generalists  are fine-tuned to target tasks using additional robot demonstrations ~\citep{gr00t_n1,psi0,dit4dit}, so deploying them on a new task requires another round of data collection. However, recordings of people walking, reaching and handling objects could expand the demonstrations available for training a humanoid generalist policy. But these recordings contain human motion, not the robot joint commands that the policy would normally learn to predict. Prior work addresses this either by having a robot mirror the person during recording, which again requires a robot for every demonstration~\citep{humanplus}, or by retargeting the human motion to the robot and using the result to supervise the policy~\citep{egomimic,hat,egoscale}.

Our observation is that the solution to the first obstacle already contains the solution to the second. Recent latent controllers \citep{sonic} map human motion and robot motion in the same latent space. For example, encoding a person walking produces a latent that commands walking on the robot when decoded by the paired controller. Although hand commands are missing in such latent controllers, a separately pretrained hand module with the same structure can cover the hands: human and robot hand encoders share a latent space, and a hand controller turns hand latents into finger commands. A generalist policy can then predict motion latents instead of joint commands, choosing where and when to move, while the frozen controllers determine how the joints must move to execute that choice. Since the generalist policy operates on the common latent space between human and robot motions, a human demonstration can supervise the learning, although the recording contains no robot joint commands. 

Built on this observation, our method,  \methodname{} uses these motion latents as the action space of a generalist policy (Figure~\ref{fig:teaser}). \methodname{} predicts chunks of body and hand latents from an egocentric image, an instruction, and proprioception. Frozen human and robot motion encoders convert demonstrations into the latents the policy learns to predict, and frozen controller converts its predictions into joint commands. This choice of action space has two benefits:
\begin{itemize}
  \item \textbf{Predicting motions, not joints.} The generalist policy selects body and hand motions rather than joint commands. The frozen controller maps the robot's measured posture and selected motion to joint commands, while a frozen hand controller converts selected hand motions into finger commands.
  \item \textbf{Human motion as supervision.} The human encoders turn recorded body and hand motion into action targets. Human images, instructions, and encoded motion train the same generalist policy as robot demonstrations, without recording a robot performing each human demonstration.
\end{itemize}

We use SONIC~\citep{sonic} for the body encoders and controller, and a separately pretrained a hand module for the hand encoders and controllers following the same structure. All encoders and controllers remain frozen during generalist-policy training. Human recordings retain their original images and instructions. The training demonstration pool contains 140.6 million frames, 93.2\% from humans. Section \ref{sec:method} describes how encoders construct targets and the generalist policy predicts them. 

On a real Unitree G1, \methodname{} achieves 100\% locomotion success and 88.6\% manipulation success without task-specific fine-tuning for either, among closing a laptop, hanging up a coat, and rotating a chair. The approach is not tied to one model, and same recipe is tested on two VLA and one WAM backbone. In addition, VioLA trained exclusively on human demonstrations shows encoded human motion can zero-shot supervise behaviors executed on the robot.

\textbf{Our contributions.} (1) \methodname{}, a generalist humanoid policy that predicts body and hand motion latents, while pretrained controllers convert those predictions into coordinated joint commands (Section~\ref{sec:method}). (2) Using human demonstrations for robot learning, converting human recordings and robot demonstrations into latent targets, creating 781 hours of training data, 93\% of which is human data. (3) Zero-shot locomotion and manipulation on a real Unitree G1 in a new environment with different object instances, without task-specific fine-tuning (Sections~\ref{sec:exp-locomotion} and~\ref{sec:exp-manipulation}), and comparisons of VLA and world-action backbones and of human and robot supervision, including real-robot locomotion learned from human demonstrations alone (Sections~\ref{sec:exp-backbones} and~\ref{sec:exp-ablations}).

%% file: sections/related_works.tex
\section{Related Work}
\label{sec:related}

\paragraph{Generalist robot policies.}
Generalist policies learn to perform many tasks from images and language instructions by training on diverse robot demonstrations. RT-2 and OpenVLA adapt pretrained vision-language models to predict actions for robot manipulation~\citep{rt2,openvla}. $\pi_0$ combines a vision-language backbone with flow matching to produce continuous actions across single-arm, dual-arm, and mobile manipulators~\citep{pi0}; $\pi_{0.5}$ extends this approach to generalization across environments through training on robot data and semantic supervision~\citep{pi05}. World-action models (WAMs) add future-video prediction: DreamZero jointly predicts video and actions, while DiT4DiT uses video-model features to condition action prediction~\citep{dreamzero,dit4dit}.

This generalist approach also extends to humanoids. GR00T and $\Psi_0$ train on diverse robot demonstrations and use task-specific teleoperation to adapt to target tasks~\citep{gr00t_n1,psi0}. $\Psi_0$ also pretrains its vision-language model to predict human wrist and hand actions, while its robot joint-command predictions are trained on robot demonstrations. Humanoid policies differ in how they pass their predictions to low-level control. Helix~02 predicts whole-body joint targets that a controller tracks~\citep{helix02}. Concurrent MotionWAM instead predicts SONIC body latents alongside direct hand or gripper commands~\citep{motionwam}. Human videos train MotionWAM's video-prediction branch, while robot demonstrations supervise its actions.

\methodname{} builds on VLA and WAM backbones, with learned motion latents as the policy outputs for both the body and hands. A pretrained body controller and hand decoder turn these predictions into joint commands. Their paired motion encoders provide targets for the same outputs from both human and robot demonstrations.

\vspace{-5pt}
\paragraph{Motion priors and hierarchical control.}
Motion imitation provides a way to learn joint coordination before learning a task. DeepMimic trains controllers to track reference movements~\citep{deepmimic}; ASE, latent-space priors, and MoConVQ learn motion representations that higher-level policies can use to control characters~\citep{ase,latent_space_priors,moconvq}. OmniH2O, HOVER, and SONIC bring motion tracking to humanoid hardware~\citep{omnih2o,hover,sonic}. In particular, SONIC pairs motion encoders with a body controller that executes their latents. LeVERB connects vision and language to motion control by first learning a latent space from rendered kinematic demonstrations, then training a controller to execute those latents~\citep{leverb}. Its hardware experiments replay latent sequences generated in simulation.

\methodname{} starts with SONIC's pretrained body encoders and controller, together with separately pretrained hand encoders and their paired robot decoders. These modules remain frozen during generalist-policy training. The generalist policy learns to predict their body and hand motion latents from images and instructions, and uses the robot's current camera observations to update its predictions during deployment.

\paragraph{Learning from human demonstrations.}
Human demonstrations can supply robot training data through physical shadowing, as in HumanPlus~\citep{humanplus}, or through reconstruction and simulation: VideoMimic reconstructs human motion and scene geometry from videos, then trains humanoid controllers to reproduce the movements~\citep{videomimic}. Other methods use recorded human motion to supervise policies directly. EgoMimic shares pose supervision between humans and robots while retaining a robot joint-action head~\citep{egomimic}. HAT and EgoVLA predict human-centric wrist and hand motion, which is converted to robot commands at deployment~\citep{hat,egovla}. EgoScale pretrains on human wrist and retargeted finger motion before adapting with human--robot and task-specific robot data~\citep{egoscale}.

Another line of work learns action labels for policy pretraining. Being-H0 pretrains a vision-language policy to predict tokenized human wrist and finger motion, then adds a robot action head during post-training~\citep{beingh0}. LAPA and WholeBodyVLA infer latent actions from video and use robot demonstrations to learn robot-command prediction~\citep{lapa,wholebodyvla}. UniT learns shared human--humanoid tokens through visual and action reconstruction~\citep{unit}. Predicting these tokens aligns policy features, while a separate prediction head produces robot-specific commands; the shared tokens themselves are not passed to a robot controller for execution.

\methodname{} uses annotated human body and hand motion to label the original human images with the same latent commands that its frozen robot decoders execute. Human demonstrations therefore supervise the generalist policy's deployed action outputs without a corresponding physical robot demonstration. Because the motion encoders and robot decoders are pretrained together, generalist-policy training does not need to learn a new mapping from human motion labels to robot joint commands.

%% file: sections/method.tex
\section{\methodname{}}
\label{sec:method}

\begin{figure}[t]
\centering
\includegraphics[width=\linewidth]{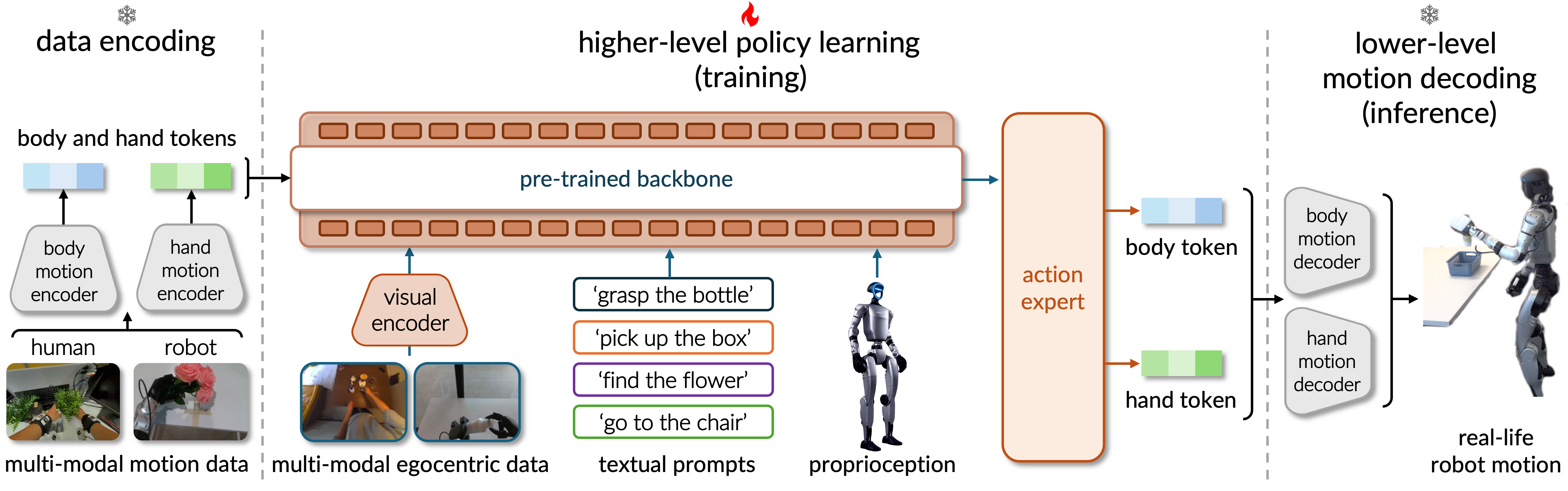}
\caption{\textbf{\methodname{} overview.}
\emph{Left:} frozen motion encoders construct body and hand action targets from human or robot demonstrations.
\emph{Center:} the generalist policy predicts motion latents from an egocentric image, an instruction, and proprioception. It is the only component updated during training.
\emph{Right:} a frozen body controller and hand decoder convert the predicted latents into robot joint commands. Human and robot encoders produce targets for these same decoders.}
\vspace{-10pt}
\label{fig:method}
\end{figure}

\methodname{} learns which body and hand motions an instruction requires while pretrained decoders produce the robot's joint commands. The key choice is to use the shared latent spaces of paired motion encoders and decoders as \methodname{}'s action space (Figure~\ref{fig:method}). Human and robot encoders map demonstrated movements into this space, supplying the targets the generalist policy learns to predict. All motion encoders and decoders are pretrained and remain frozen; training updates only the generalist policy.

\subsection{A shared latent action space}
\label{sec:motion-interface}

To close a laptop, the robot must reach for the lid and push it down while remaining balanced. A body motion encoder summarizes a short segment of recorded movement as a vector, called a motion latent. Its paired decoder is a pretrained whole-body controller: given the latent and the robot's recent state, it produces commands for the legs, torso, and arms to follow the encoded movement while maintaining balance. A separate hand encoder summarizes finger motion, and its paired decoder converts the hand latent into finger commands. \methodname{} learns to predict these body and hand latents from the scene and instruction. For the laptop task, it predicts the standing, reaching, and hand motions; the pretrained body controller and hand decoder determine the joint commands that execute them.

We use the pretrained body encoders and controller from SONIC~\citep{sonic}. Its human and robot encoders map movements into the same body latent space. Because the body controller does not command the fingers, we combine it with a hand encoder--decoder module that we pretrained separately (Appendix~\ref{sec:hand-module}). Its human and robot hand encoders likewise map motion into a common space. Each encoder summarizes a short motion window with a 64-coordinate latent. We concatenate the body and hand latents to define one action in \methodname{}'s 128-coordinate action space, $\mathbf z_t=[\mathbf z_t^{\mathrm{body}};\mathbf z_t^{\mathrm{hand}}]$.

\methodname{} predicts the body and hand latents together so that reaching and finger motion can be timed to one another. Given an egocentric image $I_t$, an instruction $\ell$, and robot proprioception $\mathbf s_t$, the policy $p_\theta$ predicts a chunk of $H$ successive actions:
\begin{equation}
    \widehat{\mathbf Z}_t \sim p_\theta(\mathbf Z_t\mid I_t,\ell,\mathbf s_t),
    \qquad
    \mathbf Z_t = [\mathbf z_t,\ldots,\mathbf z_{t+H-1}]^{\mathsf T}.
    \label{eq:chunk-policy}
\end{equation}
Each row supplies a body and hand latent for one control step; Appendix~\ref{sec:representation-details} gives the reference windows summarized by each latent. Because the paired encoders accept both human and robot motion, demonstrations from either source can supervise these predictions.

\subsection{Learning from human and robot demonstrations}
\label{sec:target-construction}
\label{sec:token-learning}

An annotated recording of a person closing a laptop contains the person's images, the instruction, and the body and hand movements used to complete it. Although it contains no robot joint commands, the human motion encoders map those movements into \methodname{}'s action space. Robot demonstrations enter the same space through the corresponding robot encoders. For demonstrator $e\in\{\mathrm{human},\mathrm{robot}\}$, the target at time $t$ is
\begin{equation}
    \mathbf z_t = \bigl[E^{e}_{\mathrm{body}}(\mathbf m_t^{\mathrm{body}};\gamma_t);\,
    E^{e}_{\mathrm{hand}}(\mathbf m_t^{\mathrm{hand}})\bigr].
    \label{eq:motion-targets}
\end{equation}
Here $\mathbf m_t^{\mathrm{body}}$ and $\mathbf m_t^{\mathrm{hand}}$ are short windows of annotated motion beginning at $t$, and $\gamma_t$ is the robot heading used to express the desired body motion. Consecutive targets form the chunk $\mathbf Z_t$.

Human data still lack one required policy input: robot proprioception. We obtain it by tracking the demonstrated motion with the frozen body controller in simulation. Body targets depend on the executing robot's current heading, so they are computed as the controller tracks the recording in simulation. The rollout also supplies robot states aligned with the human frames. Body targets for robot demonstrations use the same simulation procedure, but their policy inputs retain the recorded robot proprioception. Hand targets are encoded directly from annotated hand motion. All targets are computed before policy training; Appendix~\ref{sec:representation-details} describes the labeling procedure.

Human and robot demonstrations now have the same form: an image, an instruction, robot proprioception, and a target motion chunk. In the laptop example, the person's original camera view and instruction, paired with the simulated robot state, teach \methodname{} to predict the whole-body and hand motion. No corresponding physical robot demonstration is required. Demonstrations without hand annotations supervise only the body predictions.

Because both sources provide targets in the same action space, they train the same prediction head with the same loss. \methodname{} builds on the GR00T N1.7~\citep{gr00t_n17} backbone, adapting its action expert to predict 128 latent coordinates per step while retaining its flow-matching objective~\citep{flow_matching}. The encoders output quantized codes, but the action expert learns to predict their coordinates continuously.

\subsection{From predicted motions to continuous execution}
\label{sec:execution}

The paired decoders make the learned action space directly usable on the robot. At deployment, \methodname{} predicts a motion chunk from the robot's camera image, instruction, and proprioception. At each control step, the body controller converts the current body latent and recent proprioceptive history $\mathbf h_t$ into joint commands, while the hand decoder converts the hand latent into finger commands:
\begin{equation}
    \mathbf u_t^{\mathrm{body}} = D_{\mathrm{body}}\!\left(\widehat{\mathbf z}_t^{\mathrm{body}},\mathbf h_t\right),
    \qquad
    \mathbf u_t^{\mathrm{hand}} = D_{\mathrm{hand}}^{\mathrm{robot}}\!\left(\widehat{\mathbf z}_t^{\mathrm{hand}}\right).
    \label{eq:execution}
\end{equation}
Body and hand commands are updated at 50\,Hz. The body controller reads the robot's changing state at every update, so joint coordination continues while the generalist policy computes its next chunk. The hand decoder reconstructs finger motion from each predicted hand latent; Appendix~\ref{sec:decoder-execution} gives frame selection and smoothing details.

Continuous execution also requires consecutive chunks to agree where one replaces the other. While \methodname{} processes a new observation, the robot is already executing part of the previous prediction. An independently generated replacement can disagree with that ongoing motion and produce a jump at the switch. We therefore apply real-time chunking (RTC)~\citep{rtc} in \methodname{}'s motion-latent space: the next prediction is conditioned on the latents scheduled to execute before inference finishes. This gives the new chunk the motion it must continue. Section~\ref{sec:exp-rtc} measures the resulting continuity across chunk switches.

%% file: sections/experiments.tex
\section{Experiments}
\label{sec:experiments}

Can \methodname{} follow locomotion and manipulation instructions on a real humanoid without task-specific fine-tuning? We first evaluate task completion, then examine the motion predictions behind these behaviors. The remaining experiments test what human demonstrations contribute, which backbone learns humanoid behavior most effectively, and whether successive predictions produce continuous motion.

\subsection{Setup}
\vspace{-5pt}

\label{sec:exp-setup}

We evaluate on a Unitree G1 with five-fingered Inspire hands in our laboratory. The six locomotion and posture tasks use instructions from LeVERB; the seven manipulation tasks use instructions from Humanoid Everyday. Both corpora also contribute training demonstrations, so the evaluation tasks overlap the training task families. What is held out is the specific episodes, our laboratory environment, and the object instances; a reserved split of Humanoid Everyday is excluded from training (Appendix~\ref{sec:representation-details}). Here, zero-shot means deployment without task-specific fine-tuning. For the main comparison, each policy is tested five times per task. Success requires completing the instruction; progress also credits partial completion. We give each task equal weight and report one standard error over trials. Appendix~\ref{sec:evaluation-protocol} describes starting poses, time limits, and scoring.

The training pool combines human recordings from EgoSuite, derived from EgoStandard~\citep{egostandard}, with four robot corpora: Humanoid Everyday~\citep{humanoid_everyday}, PSI~\citep{psi0}, UnifoLM-WBT~\citep{unifolm_wbt}, and LeVERB~\citep{leverb}. The converted pool contains 140.6 million frames, 93.2\% from human demonstrations. \methodname{} uses GR00T N1.7 as its backbone and trains for 150k steps on human demonstrations, followed by 50k steps with uniform sampling across all five corpora. Appendix~\ref{sec:representation-details} gives the corpus sizes and conversion details.

We compare \methodname{} with the released GR00T N1.7~\citep{gr00t_n17} and $\Psi_0$~\citep{psi0} models to assess instruction following without task-specific adaptation. Human-only and robot-only variants of \methodname{} examine the contribution of human supervision. We also compare three \methodname{} variants built on GR00T N1.7, $\pi_{0.5}$~\citep{pi05}, and DiT4DiT~\citep{dit4dit} to assess how effectively two VLAs and one WAM learn humanoid behavior with the same training data and budget. Appendix~\ref{sec:loss-details} gives the training objectives and treatment of missing annotations.

\subsection{Zero-shot locomotion}
\label{sec:exp-locomotion}
\vspace{-5pt}

The first test is whether the policy selects the body movement specified by the instruction. The six tasks span holding a posture, raising an arm, bending, walking, sitting, and standing up from a chair. \methodname{} predicts the body and hand motions for each instruction. The frozen body controller produces the coordinated body commands, while the hand decoder produces the finger commands.

\begin{figure}[t]
\centering
\includegraphics[width=\linewidth]{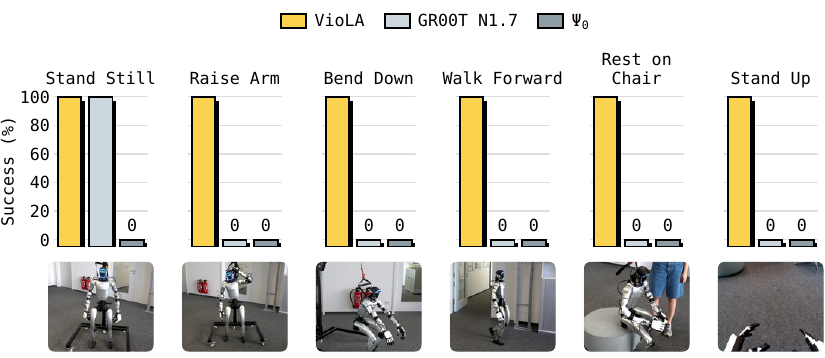}\\[4pt]
\includegraphics[width=\linewidth]{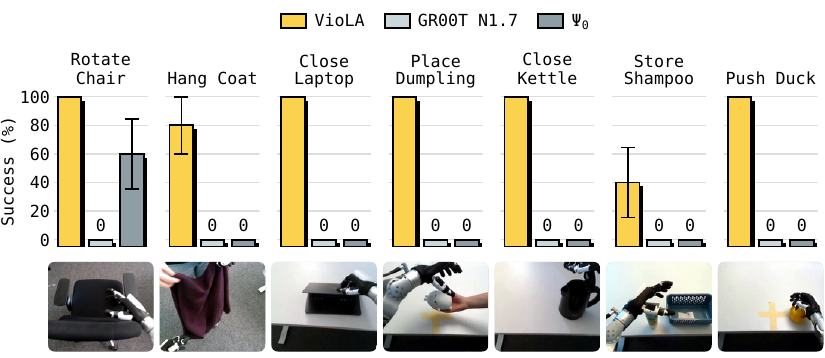}
\caption{\textbf{One policy follows locomotion and manipulation instructions without task-specific fine-tuning.} \methodname{} completes 30/30 locomotion trials (top) and 31/35 manipulation trials (bottom). Released GR00T N1.7 succeeds only at standing still; $\Psi_0$ succeeds only at chair rotation. Bars show success over five trials per task, with one standard error. Photos illustrate the tasks.}
\label{fig:main-results}
\vspace{-10pt}
\end{figure}

\methodname{} succeeds in all 30 trials, completing each task five times (Figure~\ref{fig:main-results}, top). The released GR00T N1.7 succeeds only at standing still, reaching 5/30 successes (16.7\%); $\Psi_0$ completes none (0/30). Thus, \methodname{} follows all six locomotion and posture instructions in our laboratory without task-specific fine-tuning. The Bend Down sequence in Figure~\ref{fig:rollouts} illustrates a full-body change: the robot lowers from standing into a crouched posture and holds it.

\subsection{Zero-shot manipulation}
\label{sec:exp-manipulation}
\label{sec:qualitative-rollouts}
\vspace{-5pt}

Manipulation adds a further requirement: the body must bring the hand to the object, and the fingers must act at the right time to move, grasp, or release it. We test whether the same policy can coordinate these movements through task completion, again without task-specific fine-tuning.

\methodname{} completes 31/35 trials (88.6\%; Figure~\ref{fig:main-results}, bottom). It succeeds in every trial of Rotate Chair, Close Laptop, Place Dumpling, Close Kettle, and Push Duck, and in 4/5 Hang Coat and 2/5 Store Shampoo trials. Hang Coat requires receiving a coat and folding it over the other arm. These tasks cover pushing and closing objects as well as receiving, transporting, and placing them. Released GR00T N1.7 completes none of the 35 trials. $\Psi_0$ completes three chair-rotation trials and no other manipulation task, reaching 3/35 successes (8.6\%).

\begin{figure}[t]
\centering
\includegraphics[width=\linewidth]{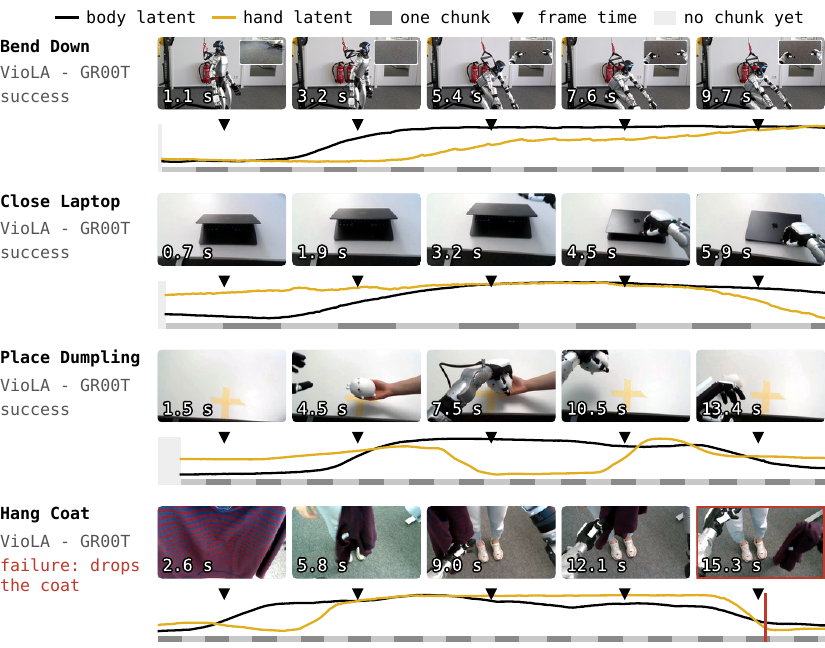}
\caption{\textbf{From body motion to object interaction.} \methodname{} bends down, closes a laptop, and places a dumpling toy on the desk (first three rows). In a failed coat execution, \methodname{} grasps the coat but then drops it (last row; failure marked in red). Black and gold curves show executed body and hand latents projected onto their first principal components (63--91\% and 47--87\% of their variance). Projections are fitted per recording and scaled separately within each row. Gray bands mark successive action chunks; triangles align image timestamps with the curves.}
\vspace{-10pt}

\label{fig:rollouts}
\end{figure}

Figure~\ref{fig:rollouts} shows how successful manipulation develops over time. To close the laptop, the robot brings its right hand to the lid and pushes it shut. To place the dumpling, it takes the toy from a person's hand with its left hand, carries it over the desk, and releases it. The accompanying latent traces show the body and hand predictions evolving across successive action chunks.

The four unsuccessful trials show difficulties in acquiring and retaining objects. In the Store Shampoo failures, the robot knocks the bottle over while reaching to grasp it. The last row of Figure~\ref{fig:rollouts} shows a failed coat execution: the robot acquires the coat but drops it before completing the folding and return instruction. These examples show why reaching the object or making an initial grasp is insufficient for manipulation success. The separate bottle-carrying task, analyzed in Appendix~\ref{sec:additional-tasks}, remains unsolved: every \methodname{} variant fails all five trials, at different stages spanning grasp timing, the transition from grasping to walking, and upright release. Including this task lowers \methodname{}'s manipulation success from 31/35 (88.6\%) to 31/40 (77.5\%).

\subsection{Predicting motions, not joints}
\label{sec:exp-tokens}
\vspace{-5pt}

How many distinct motion tokens does \methodname{} use to perform a task? We examine its body and hand predictions during standing, arm raising, walking, chair rotation, and laptop closing. Nearby latent predictions can produce nearly identical movements, so we group similar predictions and count each group as one distinct motion token. Appendix~\ref{sec:latent-analysis} explains the grouping procedure and its calibration against decoded motion. Body and hand tokens are counted separately over each complete execution.

\begin{figure}[t]
\centering
\includegraphics[width=\linewidth]{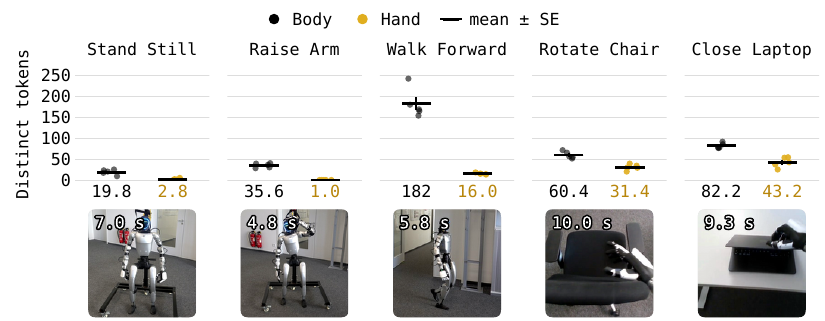}
\caption{\textbf{Complex movements have compact descriptions in motion tokens.} Each dot is the number of distinct body (black) or hand (gold) motion tokens over one complete recording, after grouping similar predictions as described in Appendix~\ref{sec:latent-analysis}. Horizontal marks show the mean and vertical whiskers one standard error across five recordings per task. Numbers below give mean counts and durations. Chair rotation and laptop closing require tens of distinct body and hand tokens over roughly ten seconds of execution.}
\label{fig:tokens}
\vspace{-5pt}
\end{figure}

Complex movements can be represented with a few motion tokens. A complete chair-rotation execution uses an average of 60 distinct body tokens and 31 hand tokens; laptop closing uses 82 body tokens and 43 hand tokens (Figure~\ref{fig:tokens}). These totals cover the entire executions, which last 10.0 and 9.3\,s on average. Standing and arm raising require fewer body tokens, about 20 and 36, while walking requires 182. The hand counts also reflect the movement: arm raising uses one hand token, whereas chair rotation and laptop closing require multiple hand motions.

These counts make the change in the prediction problem concrete. Each body token describes coordinated movement that the frozen controller turns into joint commands using the robot's changing state; each hand token specifies finger motion through the hand decoder. Complex executions therefore require a small set of distinct motion predictions, while the controllers continue to update the joints at every control step. The generalist policy learns which motions to produce and how to time them, reusing the coordination encoded by the pretrained motion modules.

\subsection{Learning from human demonstrations}
\label{sec:exp-ablations}
\vspace{-5pt}

\methodname{}'s motion action space also makes human recordings a source of action supervision. Does that supervision teach behaviors that transfer to a robot, and what does it contribute when robot demonstrations are available? We first train a variant of \methodname{} using only human demonstrations, then compare the human-then-mixed and robot-only variants. The motion encoders and decoders remain frozen in all variants.

\begin{figure}[t]
\centering
\includegraphics[width=\linewidth]{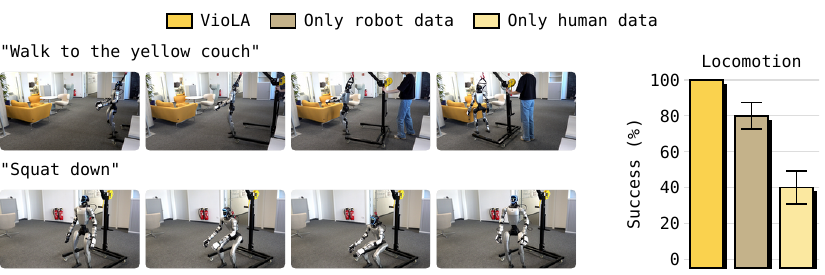}
\caption{\textbf{Human demonstrations teach robot behavior.} Left: the human-only policy follows ``walk to the yellow couch'' and ``squat down.'' Right: locomotion success with human-then-mixed training (30/30 trials), robot-only training (24/30), and human-only training (12/30). All three are \methodname{} variants with the same GR00T backbone and frozen motion modules. Error bars show one standard error over trials. The robot-only and human-then-mixed variants each receive 200k training updates.}
\label{fig:ablations}
\vspace{-15pt}
\end{figure}

The human-only \methodname{} variant succeeds in 12/30 locomotion trials (40\%): 4/5 bends, 3/5 walks, 2/5 sits, and 3/5 stands from a chair. It does not successfully follow the standing-still or arm-raising instructions. Figure~\ref{fig:ablations} also shows it walking toward a yellow couch and squatting down. These behaviors transfer without any robot demonstration in generalist-policy training. Encoded human motion therefore provides usable action supervision for real-robot behavior, although human-only training does not yet cover all six instructions reliably.

\begin{figure}[!ht]
\centering
\includegraphics[width=\linewidth]{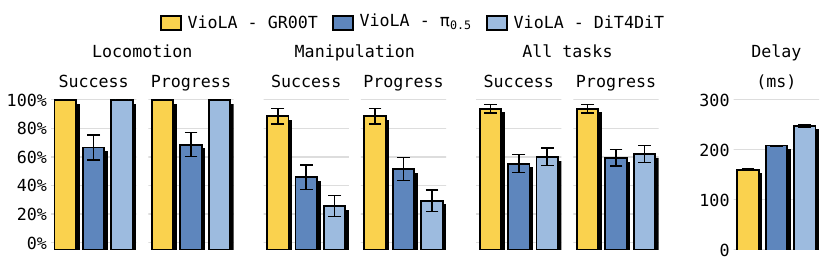}
\caption{\textbf{\methodname{} with two VLA backbones and one WAM backbone.} The variants use the same training data and budget and are evaluated without task-specific fine-tuning. Bars report success and progress over 30 locomotion trials, 35 manipulation trials, or all 65. \methodname{} reaches 100\% locomotion success with either GR00T or DiT4DiT, while GR00T performs best on manipulation. The final panel shows measured prediction delay. Error bars show one standard error over trials for performance and over recorded episodes for delay.}
\label{fig:backbones}
\vspace{-15pt}
\end{figure}

With robot demonstrations available, we compare two \methodname{} variants after 200k training steps: \methodname{} uses the human-then-mixed schedule from Section~\ref{sec:exp-setup}, while the robot-only policy uses the four robot corpora throughout. \methodname{} completes 30/30 locomotion trials, compared with 24/30 (80\%) for robot-only training. The difference comes from bending, where robot-only training fails all five trials, and one sitting trial in which it leaves the arms extended. Human-then-mixed training thus improves locomotion success by 20 percentage points at the same number of training updates.

The manipulation comparison shows a different pattern. Robot-only training completes 33/35 trials (94.3\%), compared with 31/35 (88.6\%) for \methodname{}. Robot-only training succeeds in all five shampoo trials but four kettle trials; \methodname{} succeeds in two shampoo trials and all five kettle trials. Their success counts match on the other five tasks. Human-then-mixed training therefore improves locomotion by six successful trials but completes two fewer manipulation trials. Its benefit depends on the task: it teaches body movements missing from the robot-only policy, while the robot-only variant is more successful at grasping the shampoo bottle.

\subsection{Comparing policy backbones}
\label{sec:exp-backbones}
\vspace{-5pt}

Which backbone learns humanoid behavior most effectively with \methodname{}? We compare two VLAs, GR00T N1.7 and $\pi_{0.5}$, with one WAM, DiT4DiT. All three \methodname{} variants use the same training data and training budget, predict 64 body and 64 hand coordinates per action, and execute their predictions with the same frozen motion modules. This provides a direct comparison of VLA and WAM backbones for humanoid robot learning under a common training and evaluation protocol. Each backbone retains its learning objective; DiT4DiT also predicts future video. Table~\ref{tab:backbone-contracts} gives the state inputs and chunk lengths.

\methodname{} with either GR00T or DiT4DiT succeeds in all 30 locomotion trials (Figure~\ref{fig:backbones}). The $\pi_{0.5}$ variant completes 20/30 trials (66.7\%): it succeeds at standing, arm raising, sitting, and standing up, but fails walking and full bending. Manipulation separates the integrations more strongly. \methodname{} completes 31/35 trials (88.6\%), compared with 16/35 (45.7\%) for $\pi_{0.5}$ and 9/35 (25.7\%) for DiT4DiT. \methodname{} therefore learns instruction following with both VLA and WAM backbones. GR00T gives the strongest overall results: it matches DiT4DiT on locomotion and completes substantially more manipulation trials than either alternative.

The behavior explains the gap between successful body motion and successful manipulation. All three \methodname{} variants rotate the chair in every trial. The $\pi_{0.5}$ variant also closes the laptop and places the dumpling in every trial, but it drops the coat after grasping it in four trials. DiT4DiT fails to close its thumb around the coat, and a partial dumpling trial ends in a grasp followed by a drop. Such partial completions raise manipulation progress above success, to 51.4\% for $\pi_{0.5}$ and 29.4\% for DiT4DiT. The frozen body controller and hand decoder are shared across the three variants. The differences arise in the generalist policy's predictions: successful manipulation requires the right reach, grasp, and release sequence through the end of the task.

For \methodname{}, prediction delay is 160.2\,ms with GR00T, 208.6\,ms with $\pi_{0.5}$, and 248.5\,ms with DiT4DiT. These measurements span the recorded deployments summarized in Appendix~\ref{sec:recording-timing}. At 50\,Hz, each prediction takes roughly eight to twelve control steps, during which the robot continues to move. This delay makes the transition from the ongoing motion to the next prediction an essential part of execution.

\subsection{Continuity across motion chunks}
\label{sec:exp-rtc}
\vspace{-5pt}

Does replanning preserve the motion already in progress? While a new chunk is being predicted, the robot executes part of the previous chunk. The two predictions can disagree at the point where the new one takes over. We test whether real-time chunking (RTC)~\citep{rtc}, which conditions the next prediction on the latents scheduled for execution during inference, reduces this disagreement in motion space.

\begin{figure}[t]
\centering
\includegraphics[width=\linewidth]{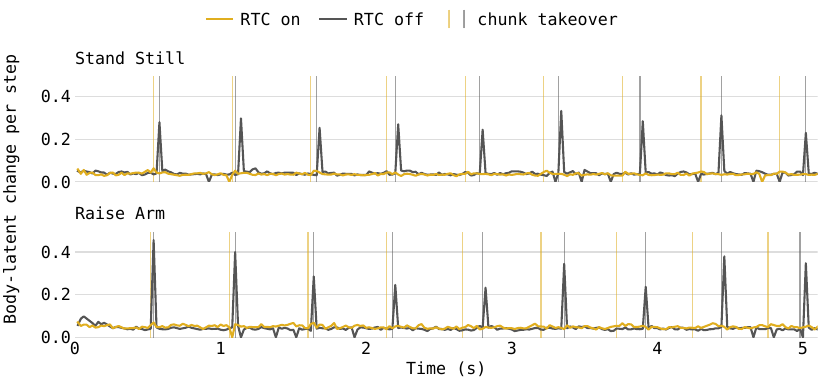}
\vspace{-20pt}
\caption{\textbf{RTC reduces jumps when a new motion chunk takes over.} Each row compares \methodname{} recordings with RTC on (gold) and off (gray). Curves show the L2 change in the 64-coordinate body latent between consecutive 20\,ms control steps. Vertical lines mark chunk switches in the corresponding condition's color. RTC-off traces show sharp spikes near these switches, whereas RTC-on traces remain close to their between-switch variation. Each recording is aligned to the start of policy execution and shown for 5.1\,s; the initial idle-to-policy transition is excluded.}
\label{fig:rtc}
\vspace{-15pt}
\end{figure}

We compare \methodname{} with RTC enabled and disabled on Stand Still and Raise Arm. The first tests consistency while holding a posture; the second tests continuity while the requested posture changes. Figure~\ref{fig:rtc} measures the L2 change in the body latent between consecutive 20\,ms control steps, $\Delta_t=\|\mathbf z_t^{\mathrm{body}}-\mathbf z_{t-1}^{\mathrm{body}}\|_2$. Vertical markers show when execution switches to a new chunk. A spike at a marker indicates an abrupt change in the predicted body latent when the chunk is replaced.

Without RTC, sharp spikes recur near chunk switches, even for Stand Still. With RTC, the changes at these switches stay close to the variation between switches. Over the displayed 5.1\,s windows, the largest step change decreases from 0.332 to 0.064 for Stand Still and from 0.455 to 0.070 for Raise Arm. RTC therefore reduces discontinuities in the predicted body latents when consecutive chunks are joined in these recordings. Appendix~\ref{sec:rtc-measurement} gives the measurement and timing details.

%% file: sections/conclusion.tex
\section{Limitations and Conclusion}
\label{sec:conclusion}

\paragraph{Limitations.}
\methodname{} reuses pretrained body and hand modules, so its capabilities depend on the movements those modules can execute. The hand decoder produces finger-position commands without contact-force feedback. The manipulation trials show remaining difficulties in reaching to grasp objects and retaining them through task completion. Incorporating contact feedback into hand control is one direction for improving object interaction.

Learning from human recordings requires human body and hand motion annotations. Extending \methodname{} to additional video therefore depends on obtaining reliable estimates of that motion. Human supervision also does not improve every behavior under the current training schedule: human-then-mixed training improves locomotion, but robot-only training achieves higher manipulation success. How to retain the locomotion benefit while improving object interaction remains an open question.

The main evaluation covers thirteen tasks on one Unitree G1, with five trials per task. Broader testing across tasks, environments, and robot embodiments is needed to assess how reliably \methodname{} performs beyond this evaluation.

\paragraph{Conclusion.}
\methodname{} is a generalist humanoid policy that learns from human and robot demonstrations to predict body and hand movements. Pretrained motion encoders convert the demonstrations into latent targets; \methodname{} predicts those latents from images, instructions, and proprioception. A frozen body controller turns the body predictions into coordinated joint commands while maintaining balance, and a frozen hand decoder produces the finger commands. Human recordings can therefore teach \methodname{} which movements an instruction requires without a robot performing each demonstration.

On a real Unitree G1, \methodname{} achieves 100\% locomotion success and 88.6\% manipulation success without task-specific fine-tuning. Its complex movements are represented with a few motion tokens. Comparing two VLA backbones and one WAM backbone using the same training data and budget shows that \methodname{} performs best with GR00T, particularly for manipulation. Training the generalist policy on human demonstrations alone still yields 40\% locomotion success on the real robot.

%% file: sections/implementation.tex
\section{Implementation and Additional Analyses}
\label{sec:implementation}

\subsection{Motion representation and data conversion}
\label{sec:representation-details}

\paragraph{Token coordinates and reference windows.}
Each encoder produces a 64-coordinate motion code, formed by flattening two 32-dimensional finite scalar quantization (FSQ) codes~\citep{fsq}. Coordinates take values in $\{k/16:k=-16,\ldots,15\}$. Body and hand codes use this same format, but have separate encoders and decoders. A code summarizes a reference window beginning at the current frame. At 50\,Hz, the SONIC v1.1 robot encoder and the hand encoders read ten frames at offsets $0,5,\ldots,45$; the human body encoder reads ten consecutive frames at offsets $0,1,\ldots,9$. Windows extending beyond the demonstration repeat its final frame. This reference window determines what one code represents; the policy's chunk length $H$ determines how many successive codes it predicts.

\paragraph{Motion targets.}
Equation~\ref{eq:motion-targets} concatenates the body and hand encoder outputs. In this implementation, $\mathbf z_t\in\mathcal G^{128}$ with $\mathcal G=\{k/16:k=-16,\ldots,15\}$. The human body encoder reads SMPL motion~\citep{smpl}; human hand motion is obtained by fitting MANO~\citep{mano} to annotated keypoints. Each training example pairs the chunk with an image, instruction, and robot state:
\begin{equation}
    \bigl(I_t,\;\ell,\;\mathbf s_t,\;\mathbf Z_t\bigr),
    \qquad
    \mathbf Z_t = [\mathbf z_t,\ldots,\mathbf z_{t+H-1}]^{\mathsf T},
    \label{eq:training-example}
\end{equation}
Robot examples use recorded proprioception, while human examples use simulated robot states. The body targets for both sources are constructed through the simulation procedure below. This converts annotated human motion into latent targets without optimizing a separate mapping to robot joints for each recording.

\paragraph{Closed-loop construction of body targets.}
Body targets for both human and robot demonstrations are generated by tracking the reference motion with the frozen controller in simulation. The body encoder expresses the desired motion relative to the executing robot's heading, which can differ from the demonstrator's heading. For example, if the robot turns toward a table more slowly than the person in the recording, the next latent must encode the turn that remains. Using the person's heading throughout would assume that the robot follows the reference exactly. We therefore compute the targets in closed loop. At each simulation step, the encoder uses the simulated robot's current heading to compute a code, the controller decodes it into joint commands, and the simulator advances:
\begin{equation}
    \mathbf z_t^{\mathrm{body}} = E^{e}_{\mathrm{body}}\!\left(\mathbf m_t^{\mathrm{body}};\,\gamma(\mathbf s_t^{\mathrm{sim}})\right),
    \qquad
    \mathbf s_{t+1}^{\mathrm{sim}} = \mathrm{Sim}\!\left(\mathbf s_t^{\mathrm{sim}},\,D_{\mathrm{body}}(\mathbf z_t^{\mathrm{body}},\mathbf h_t^{\mathrm{sim}})\right).
    \label{eq:closed-loop-encoding}
\end{equation}
At time $t$, the image $I_t$, simulated state $\mathbf s_t^{\mathrm{sim}}$, and encoded target $\mathbf z_t^{\mathrm{body}}$ are paired before the simulator advances to $\mathbf s_{t+1}^{\mathrm{sim}}$. Human examples use this simulated state as body proprioception; robot examples retain their recorded proprioception.

Hand-state inputs use the values stored in the data. For robot demonstrations, hand targets come from commanded finger motion where available, while hand proprioception comes from measured motion. This distinction matters during grasping: the commanded fingers can close farther than the object permits them to move. Examples without hand annotations contribute only to the body loss (Equation~\ref{eq:motion-loss}). Motion targets are computed before generalist-policy training; training does not run the simulator or update either motion module.

\paragraph{Body-controller configuration.}
SONIC v1.1 uses a 1,751-coordinate encoder input containing the reference motion in the simulated robot's heading frame. The body decoder receives the current 64-coordinate code and ten frames of proprioceptive history: joint offsets, joint velocities, base angular velocity, projected gravity, and previous actions. Together these form 994 input coordinates. The decoder's 29 outputs are converted to joint-position targets using the controller's default pose and action scale. Control runs at 50\,Hz, with four 200\,Hz physics steps per update.

\paragraph{Available converted data.}
The training pool combines human recordings from EgoSuite, derived from EgoStandard~\citep{egostandard}, with four robot corpora: Humanoid Everyday~\citep{humanoid_everyday}, PSI~\citep{psi0}, UnifoLM-WBT~\citep{unifolm_wbt}, and LeVERB~\citep{leverb}.
Table~\ref{tab:converted-data} lists the retained demonstrations before the training/evaluation split. The converted human pool contains 20,680 of the 26,382 source episodes; 5,702 are excluded during conversion and filtering. The unified loader reserves 338 Humanoid Everyday episodes for evaluation. It also excludes episodes too short to supply a complete action chunk; at $H=50$, 46 of the 3,555 LeVERB episodes provide no training window. Humanoid Everyday and UnifoLM-WBT supply commanded hand motion, whereas PSI supplies measured hand motion because commanded trajectories are unavailable. Human hand targets are derived from fitted MANO trajectories. LeVERB supplies body targets only.

\begin{table}[t]
\centering
\small
\begin{tabular}{lrrrl}
\hline
Source & Episodes & Frames & Hours & Targets \\
\hline
Humanoid Everyday & 4,064 & 2,962,764 & 16.46 & Body and hands \\
PSI & 835 & 1,431,454 & 7.95 & Body and hands \\
UnifoLM-WBT & 2,854 & 4,454,240 & 24.75 & Body and hands \\
LeVERB & 3,555 & 713,279 & 3.96 & Body only \\
EgoSuite & 20,680 & 131,010,575 & 727.84 & Body and hands \\
\hline
Total & 31,988 & 140,572,312 & 780.96 & \\
\hline
\end{tabular}
\caption{\textbf{Converted demonstration pool.} Humanoid Everyday, PSI, and UnifoLM-WBT contain real-G1 demonstrations; LeVERB contains simulated G1 demonstrations, and EgoSuite contains human recordings. All are converted to 50\,Hz, with hand targets where available. Hours are frames divided by 50\,Hz. Human data are 64.6\% of episodes and 93.2\% of frames. Sampling weights, not corpus size, determine exposure to each source during training.}
\label{tab:converted-data}
\end{table}

\begin{figure}[t]
\centering
\includegraphics{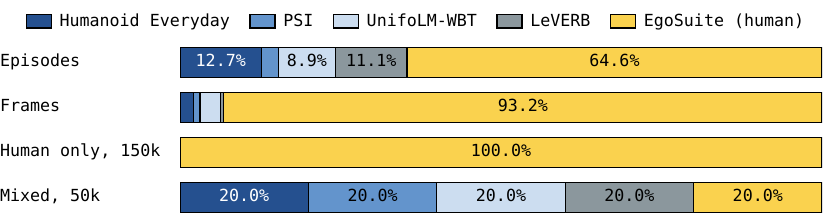}
\caption{\textbf{Dataset size and \methodname{} training schedule.} The first two bars show each source's share of episodes and frames in the converted pool. The last two show sampling probabilities during 150k steps of human-only training and 50k steps of uniform sampling over all five corpora. Human recordings supply 93.2\% of available frames; the training schedule determines how often they are sampled.}
\label{fig:data-share}
\end{figure}

\methodname{} trains for 150k steps on human demonstrations, followed by 50k steps with sampling probability 0.20 for each of the five corpora (Figure~\ref{fig:data-share}). The robot-only variants use the four robot corpora throughout and are evaluated after 50k and 200k steps. The human-only variant uses human demonstrations throughout. No variant receives task-specific adaptation. With a constant batch size across stages, the \methodname{} schedule assigns an expected 80\% of training examples to human data, distinct from their 93.2\% share of the available frames.

\subsection{Hand motion interface}
\label{sec:hand-module}

We pretrain a paired hand encoder--decoder module that maps human hand motion and robot hand motion into a shared 64-coordinate space. Its Dex3 and Inspire decoders translate these codes into finger trajectories for the corresponding hardware. All hand encoders and decoders remain frozen during generalist-policy training. The policy learns to predict their codes; it does not learn the hand prior itself.

\subsection{Backbone and action-expert instantiations}

The GR00T, $\pi_{0.5}$, and DiT4DiT variants of \methodname{} use the same training data and training budget. Each predicts 128 coordinates per action step: 64 for the body and 64 for the hands. \methodname{} with GR00T predicts $H=40$ successive steps; the other variants use $H=50$ (Table~\ref{tab:backbone-contracts}). State inputs differ by backbone. The $\pi_{0.5}$ configuration uses 29 body joint offsets, three gravity coordinates, 14 current finger positions expressed in the Dex3 representation, and a 64-coordinate hand code delayed by 50 ticks. The one-second delay prevents the code's reference window from exposing future targets. During training, this delayed code is zeroed with probability 0.5; it can also be zeroed during deployment. Current finger state is reconstructed from measured hand motion, using frame zero of the hand decoder when conversion is needed.

\begin{table}[t]
\centering
\small
\begin{tabular}{lrll}
\hline
Backbone family & $H$ & State conditioning & Motion-target scaling \\
\hline
$\pi_{0.5}$ & 50 & 110 values, tokenized & Identity \\
GR00T N1.7 & 40 & 46 values, state encoder & Identity \\
DiT4DiT & 50 & 32 values, state encoder & Identity \\
\hline
\end{tabular}
\caption{\textbf{Backbone-specific interfaces for \methodname{}.} All variants predict the same 128 body and hand coordinates without target normalization. Chunk length and state conditioning follow each backbone.}
\label{tab:backbone-contracts}
\end{table}

GR00T N1.7~\citep{gr00t_n17} combines a Cosmos-Reason2 vision-language backbone with a flow-matching action transformer. Its state encoder receives 29 body joint offsets, 14 finger positions, and three gravity coordinates, with state dropout of 0.2. Human and robot examples use the same action expert and embodiment tag. All methods predict native FSQ coordinates without action-target normalization. State inputs are normalized with statistics fitted on training data. $\pi_{0.5}$ discretizes the normalized values into 256 bins for its input prompt, whereas GR00T and DiT4DiT project them into learned features.

DiT4DiT uses Cosmos-Predict2.5-2B, with features from layer 17 conditioning the action transformer. During training, the video model receives supervision from the current frame and four future frames at offsets $12,24,36,48$ on the 50\,Hz timeline. The objective adds video and motion flow-matching losses, and gradients from the action loss also update the video backbone. At deployment, only the current observation is available; future video features are predicted by the model.

\subsection{Flow-matching conventions and masks}
\label{sec:loss-details}

The action expert $v_\psi$ is conditioned on backbone features and robot state, collectively denoted $\mathbf c_t$. Flow matching mixes each target with noise and trains the expert to predict the difference. Under the convention below, inference starts from noise at $\tau=1$ and integrates the learned velocity backward in time toward a clean action chunk at $\tau=0$. Encoder codes are treated as continuous coordinates in this regression.

For a target chunk $\mathbf Z_t$, noise $\boldsymbol\epsilon\sim\mathcal N(\mathbf 0,\mathbf I)$ of the same shape, and an interpolation time $\tau$,
\begin{equation}
    \mathbf X_{t,\tau} = (1-\tau)\mathbf Z_t + \tau\boldsymbol\epsilon,
    \qquad
    \mathbf U_t = \boldsymbol\epsilon-\mathbf Z_t,
    \label{eq:flow-path}
\end{equation}
and the action expert $v_\psi$ is trained to output $\mathbf U_t$ from $\mathbf X_{t,\tau}$:
\begin{equation}
    \mathcal L_{\mathrm{motion}} =
    \mathbb E_{\boldsymbol\epsilon,\tau}\!\left[
    \sum_{n\in\mathcal B}\sum_{k=0}^{H-1}\sum_{j=1}^{128}
    w_{nkj}\bigl(v_\psi(\mathbf X_{t,\tau},\tau\mid\mathbf c_t)_{n,kj}-U_{n,kj}\bigr)^2
    \right],
    \label{eq:motion-loss}
\end{equation}
where $n$ indexes the examples in a batch $\mathcal B$, each with its own target chunk, noise, and conditioning, $k$ the step within the chunk, and $j$ the latent coordinate. The weight $w_{nkj}$ normalizes the loss and is zero for coordinates without a target, such as the hand coordinates of a recording without hand annotations. The reductions below specify each backbone's normalization. Human and robot examples pass through the same action expert and the same loss; no explicit human--robot source label is supplied to the action expert. The motion encoders and decoders remain frozen.

DiT4DiT is a world-action model: it is trained to predict the future video frames $\mathbf V_t$ that follow the current image, as well as the actions. We keep that objective. Its video branch uses the same squared velocity-prediction principle, with its own video representation and loss reduction. The video and motion losses are summed:
\begin{equation}
    \mathcal L_{\mathrm{WAM}} = \mathcal L_{\mathrm{video}}(\mathbf V_t\mid I_t,\ell) + \mathcal L_{\mathrm{motion}}(\mathbf Z_t\mid I_t,\ell,\mathbf s_t).
    \label{eq:wam-loss}
\end{equation}
For a human clip, the recorded future frames supervise video prediction, and the encoded body and hand motion supervises action prediction.

The loss excludes invalid frames and missing hand annotations. Let $N$ denote batch size and $m_{nkj}\in\{0,1\}$ indicate whether coordinate $j$ at chunk step $k$ in example $n$ has a valid target. For $\pi_{0.5}$, Equation~\ref{eq:motion-loss} uses
\begin{equation}
    w_{nkj}=\frac{m_{nkj}}{NH\max(1,\sum_{j'}m_{nkj'})}.
    \label{eq:pi-loss-weights}
\end{equation}
Thus each step first averages over its annotated coordinates. GR00T and DiT4DiT instead average over all valid coordinates in the batch:
\begin{equation}
    w_{nkj}=\frac{m_{nkj}}{\sum_{n',k',j'}m_{n'k'j'}}.
    \label{eq:global-loss-weights}
\end{equation}
GR00T adds $10^{-6}$ to the denominator, and DiT4DiT requires at least one valid coordinate. For fully annotated batches, the reductions are equivalent up to the numerical offset. With missing hand annotations, the first gives equal weight to action steps, while the second gives equal weight to annotated coordinates.

Equation~\ref{eq:flow-path} uses $\tau=1$ for noise and $\tau=0$ for clean targets. The $\pi_{0.5}$ implementation samples $\tau=0.999b+0.001$, with $b\sim\operatorname{Beta}(1.5,1)$. DiT4DiT samples $\tau=b/0.999$ with the same beta parameters and discretizes time into 1,000 conditioning buckets. GR00T uses the opposite time direction: its implementation samples $r=0.999(1-b)$, forms $(1-r)\boldsymbol\epsilon+r\mathbf Z_t$, and regresses $\mathbf Z_t-\boldsymbol\epsilon$. Setting $\tau=1-r$ and reversing the velocity sign gives the convention of Equation~\ref{eq:flow-path}. Each backbone retains its own time sampler and numerical solver settings.

\subsection{Decoder execution}
\label{sec:decoder-execution}

The body controller updates its proprioceptive history and joint commands every 20\,ms, while the generalist policy computes its next chunk. The hand decoder reconstructs ten frames from each hand latent. Execution selects frame index five (zero-based), applies a five-tick moving average, and converts the result to the hand's joint commands. Frame zero is used separately to reconstruct current hand state.

The Unitree G1 uses five-fingered Inspire hands, with six actuated joints per hand. Predicted latents pass through the Inspire decoder to produce 12 finger targets. For policies trained with a 14-value Dex3 hand state, measured Inspire motion is encoded and then reconstructed with the Dex3 decoder. This conversion changes the state representation presented to the policy; the execution targets still use the Inspire decoder.

Although encoder targets lie on the FSQ grid, the action experts predict continuous coordinates. The $\pi_{0.5}$ execution path clips predictions to $[-1,1]$ and passes them directly to the decoders without rounding. The hand decoder is selected for the robot's hardware before converting its output to finger commands.

\paragraph{Chunk replacement and RTC.}
While the decoders execute a chunk, the generalist policy uses a new image and robot state to predict its replacement. If the predictions disagree where the new chunk takes over, the commanded motion can change abruptly. RTC operates on motion latents, conditioning the next prediction on the part of the current chunk that will execute before inference finishes. Keeping these latents fixed helps the new chunk continue the current motion. Section~\ref{sec:exp-rtc} compares transitions with and without RTC.

\subsection{Evaluation protocol}
\label{sec:evaluation-protocol}

Instructions follow the task descriptions in LeVERB and Humanoid Everyday. All tasks are evaluated in our laboratory with locally available objects rather than the physical objects in the datasets. Each policy is tested five times per task. Trials begin from an idle standing pose, except Stand Up, which begins seated on a chair. Locomotion trials allow 5\,s, except Bend Down, which allows 10\,s. Manipulation trials have no fixed time limit. For \methodname{}, a frozen SONIC v1.1 controller and a hand decoder execute motion predictions at 50\,Hz.

Success requires completing the instruction. Locomotion trials test standing, raising an arm, bending, walking forward, remaining seated, or rising from a chair. Standing up and then walking still counts as success; sitting and immediately standing again does not. Manipulation trials require rotating the chair, folding and returning the coat, closing the laptop or kettle lid, placing the dumpling on the desk, putting the shampoo in the container, or pushing the duck to the desk's center. Progress assigns intermediate scores to incomplete executions. The robot-only policy receives 0.5 for sitting with its arms still extended instead of lowering them. A DiT4DiT kettle trial receives 0.8 after slowly closing the lid but continuing the closing motion without returning the arm to idle. Thus, the manipulation scores also account for how the robot concludes the action, despite the absence of a fixed timeout. Only a score of one counts as success. We average scores over five trials per task, giving each task equal weight; error bars show one standard error.

\subsection{Failures in grasping and object transport}
\label{sec:additional-tasks}

\paragraph{Bottle carrying.}
Bottle carrying requires the robot to take a bottle from a person with its left hand, walk to a table, and place it upright. \methodname{} fails all five trials. It approaches the table but releases the bottle while tilted, causing it to fall. The $\pi_{0.5}$ and DiT4DiT variants of \methodname{}, and the robot-only GR00T variant trained for 200k steps, also fail all five trials. This task is analyzed separately from the seven manipulation tasks in the main figure; including it reduces \methodname{}'s manipulation success from 31/35 (88.6\%) to 31/40 (77.5\%).

The \methodname{} variants fail at different stages. The robot-only variant grasps the bottle but does not start walking toward the table. The $\pi_{0.5}$ variant repeatedly reaches toward it but rarely closes its hand. The DiT4DiT variant closes its hand before reaching the bottle. Together, these failures show that grasp timing, the transition from grasping to walking, and stable release remain difficult.

\begin{figure}[t]
\centering
\includegraphics{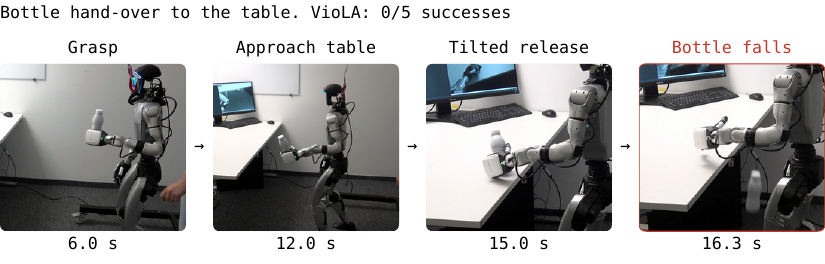}
\caption{\textbf{Bottle carrying fails at release.} One \methodname{} trial from a third-person camera. The robot takes the bottle from a person, walks to the table, and opens its hand while the bottle is tilted; the bottle then falls off the table. Times are seconds from the start of the clip. The task scores 0/5 and is reported separately from the seven manipulation tasks in Figure~\ref{fig:main-results}.}
\label{fig:bottle-failure}
\end{figure}

\paragraph{Other manipulation failures.}
\methodname{} also fails three of five shampoo trials by knocking the bottle over during the reach, and one of five coat trials. In the failed coat execution shown in Figure~\ref{fig:rollouts}, the robot acquires the coat but drops it before completing the instruction. These cases distinguish reaching the object from maintaining control of it through placement or handover.

\subsection{Execution timing}
\label{sec:recording-timing}

Timing is measured over 161 episodes of \methodname{} with GR00T, 83 with $\pi_{0.5}$, and 57 with DiT4DiT, including recordings beyond the five scored trials per task. For each episode, prediction delay is the mean time from requesting a chunk to receiving it. The replanning interval is the mean time between consecutive chunk arrivals, measured from their frame indices at 50\,Hz. Table~\ref{tab:timing} averages these episode-level measurements, giving each episode equal weight.

\begin{table}[t]
\centering
\small
\begin{tabular}{lrrcc}
\hline
Backbone & Episodes & Chunks & Delay (ms) & Replan interval (ms) \\
\hline
GR00T & 161 & 3,871 & $160.2\pm1.8$ & $559.9\pm1.8$ \\
$\pi_{0.5}$ & 83 & 2,554 & $208.6\pm0.6$ & $608.4\pm0.6$ \\
DiT4DiT & 57 & 1,035 & $248.5\pm2.5$ & $648.0\pm2.4$ \\
\hline
\end{tabular}
\caption{\textbf{Measured chunk timing.} Mean $\pm$ one standard error across recorded episodes. Delay measures the time from requesting to receiving a prediction; the replanning interval measures time between successive chunk arrivals. The episode counts exceed the scored-trial counts because the recording archive includes additional runs.}
\label{tab:timing}
\end{table}

\subsection{Measuring chunk-transition continuity}
\label{sec:rtc-measurement}

For each task and RTC condition, Figure~\ref{fig:rtc} uses the first recording in chronological order with at least four seconds of policy execution. All four traces are cropped to 5.1\,s, the shortest selected duration rounded down to a tenth of a second. This selection does not use latent-change magnitude. The metric is computed directly from the logged, continuous body latents, without grid rounding or projection. Consecutive steps are identified by recording timestamps separated by 20\,ms, with a 1\,ms tolerance. Differences across timestamp gaps are excluded; the four selected recordings contain no such gaps. Repeated or skipped policy-counter values do not exclude a sample when its recording timestamp remains consecutive. The first policy command is excluded because its preceding command belongs to the idle pose token.

Each chunk-switch marker indicates when execution begins using a new prediction. It is placed at the timestamp of the first recorded sample whose policy counter reaches the new chunk's logged takeover index. Elapsed time is measured relative to the first policy sample.

The selected recordings run at 50\,Hz. Mean prediction delays are 138.9\,ms with RTC and 154.1\,ms without RTC for Stand Still, and 130.0\,ms and 155.3\,ms for Raise Arm. Corresponding mean intervals between chunk switches are approximately 537 and 552\,ms for Stand Still, and 529 and 553\,ms for Raise Arm, measured from the timestamp-aligned takeover markers. Thus the traces compare recorded executions with slightly different timing, rather than imposing identical inference delays. The reported maxima describe these four examples, not averages over repeated trials.

\subsection{Motion token counts and calibration}
\label{sec:latent-analysis}
\paragraph{Counting distinct motion tokens.}
We count the distinct representatives needed to describe the policy's body and hand predictions over a task execution. Each 64-coordinate latent $\mathbf z$ is mapped to integer grid coordinates solely to measure grouping distance:
\begin{equation}
    \mathbf q=\mathrm{clip}\bigl(\mathrm{round}(16\mathbf z),-16,15\bigr),
    \qquad
    d(\mathbf q,\mathbf q')=\sum_{j=1}^{64}|q_j-q'_j|.
    \label{eq:grouping-distance}
\end{equation}
We scan each recording in time order, retaining the first prediction. A subsequent prediction is assigned to its nearest retained representative if the distance is at most the threshold $\rho$; otherwise, it becomes a new representative. Ties select the earliest representative. Representatives are original predicted vectors, not averages or grid-rounded vectors. Even $\rho=0$ can group unequal continuous predictions that fall in the same grid cell. Body and hand predictions are grouped separately, with a new set of representatives for each execution. We report counts per execution, with duration for context, and average counts across executions of each task.

\paragraph{Decoded-pose calibration.}
Calibration uses 84 episodes: 32 chair-rotation and 32 laptop-closing demonstrations, plus 10 policy recordings each of Place Dumpling and Push Duck. The demonstration latents are encoder targets; the policy latents are continuous predictions. A further 21 episodes are held out for validation, with 8, 8, 3, and 2 episodes from the respective tasks. We test body thresholds in $\{0,1,2,4,8,16,32,64\}$ and extend the hand grid with $\{96,128,192,256,384,512\}$. At each threshold, we group the latents and compare the decoded poses of each original latent and its representative at the same robot state. Body decoder outputs are converted to joint-position targets, and forward kinematics gives link positions for the comparison. Body error is the mean root-relative position difference over the 14 links used by SONIC's MPJPE-L metric. Hand error is the mean position difference over six Dex3 distal-finger link centers of mass, used as fingertip proxies, with the decoder evaluated at frame zero. Each episode's error is the 95th percentile of its framewise mean point errors. Invalid frames and frames at or after a simulated fall are excluded.

\paragraph{Tracking-error references.}
We use tracking accuracy to set a physical reference scale for the grouping error. For the body, the reference is SONIC's reported real-world MPJPE-L of 25.7\,mm (2.57\,cm) over 14 links~\citep{sonic}. For the hands, we measure the discrepancy between commanded and observed finger joint poses in 83 Humanoid Everyday episodes (chair rotation, laptop closing, and chair-alt). Forward kinematics gives a mean error of 1.44\,cm over six fingertip proxies and 32,516 frames. This hand reference uses the same points as the grouping analysis. These values are tracking-error reference budgets, not task-success tolerances or bounds on the effect of replacing a complete motion sequence.

\paragraph{Threshold selection.}
For each modality, we find the largest tested threshold for which at least 95\% of calibration episodes have their episode error within the tracking-error reference budget. This rule permits thresholds of 16 for the body and 96 for the hands; requiring every calibration episode to pass gives the same values. Episodes receive equal weight, and none are removed as outliers.

We report finer thresholds of 8 for the body and 32 for the hands because averaging over links or fingertips can hide larger deviations at individual points. Reducing the body threshold from 16 to 8 lowers the mean episode-level wrist p95 from 5.09 to 3.03\,cm on calibration episodes. Reducing the hand threshold from 96 to 32 lowers the worst-finger p95 from 1.08 to 0.31\,cm for chair rotation and from 1.25 to 0.31\,cm for laptop closing. At these finer thresholds, both modalities satisfy their mean-error budgets in all 84 calibration episodes, all 21 held-out episodes, and three additional chair-alt transfer episodes. Across the calibration and held-out episodes, the largest episode-level 95th-percentile mean error is 1.234\,cm for body links and 0.294\,cm for fingertip proxies. Figure~\ref{fig:tokens} reports counts over five recordings per task at these thresholds.

\begin{figure}[tp]
\centering
\includegraphics[width=\linewidth,height=0.8\textheight,keepaspectratio]{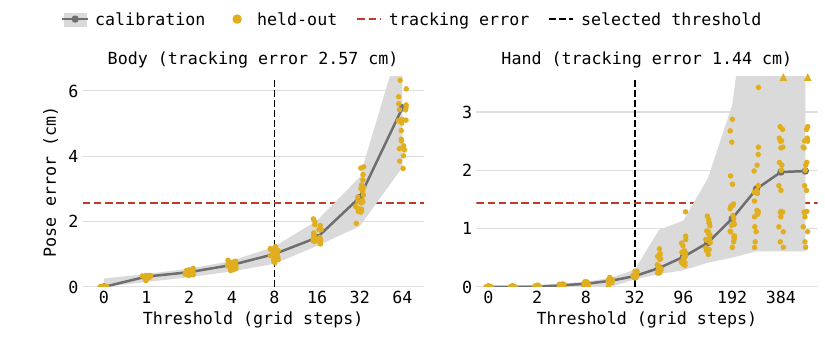}
\caption{\textbf{Threshold calibration.} Each episode is summarized by the 95th percentile of framewise error, averaged over 14 body links or six fingertip proxies. The gray line shows the median of the calibration episodes' errors; shading spans their minimum and maximum. Gold dots show held-out episodes; triangles mark values above the displayed range. Red horizontal lines mark the tracking-error references (2.57\,cm body, 1.44\,cm hand). Vertical dashed lines mark the reported thresholds of 8 and 32, chosen finer than the permitted 16 and 96 to reduce wrist and individual-finger deviations.}
\label{fig:token-calibration}

\vspace{1em}

\includegraphics[width=\linewidth,height=0.8\textheight,keepaspectratio]{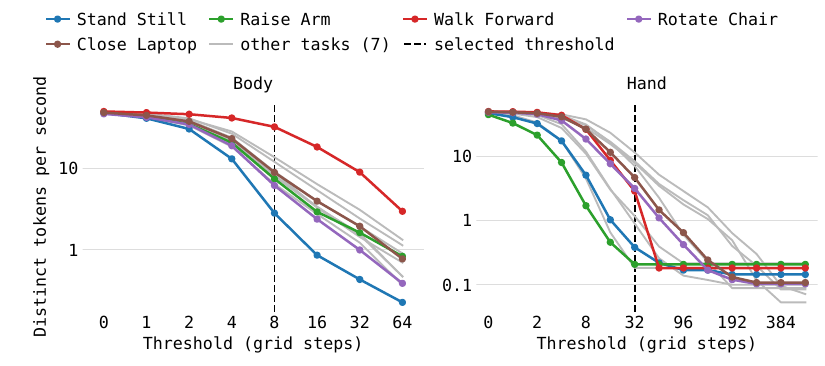}
\caption{\textbf{Token counts vs.\ threshold.} Curves show how grouping distance changes the number of distinct body and hand representatives. Each recording's count is divided by its duration before averaging within each task; the vertical axis is logarithmic. Colored curves identify the five tasks in Figure~\ref{fig:tokens}; gray curves show seven additional tasks. Dashed lines mark the reported body and hand thresholds of 8 and 32.}
\label{fig:token-thresholds}
\end{figure}

Figures~\ref{fig:token-calibration} and~\ref{fig:token-thresholds} report the error curves and sensitivity of counts to the grouping threshold. The robot executes the original continuous predictions; the analysis describes them at a resolution referenced to tracking accuracy. It measures local differences in decoded pose, not accumulated execution error. Hand calibration uses Dex3 fingertip proxies rather than the deployed Inspire hand trajectory.

